\documentclass{article}
\usepackage{spconf,amsmath,graphicx}
\usepackage{color, colortbl}
\usepackage{subfig}
\definecolor{lgray}{rgb}{0.88,0.88,0.88}
\usepackage{pgfplots,wrapfig}
\usetikzlibrary {backgrounds} 

\title{Real-time Traffic Object Detection for Autonomous Driving}
\name{Abdul Hannan Khan$^{1,2}$, Syed Tahseen Raza Rizvi$^{2}$, Andreas Dengel$^{1,2}$}
\address{
Department of Computer Science, RPTU Kaiserslautern-Landau$^{1}$,\\
German Research Center for Artificial Intelligence (DFKI GmbH)$^{2}$,\\
67663 Kaiserslautern, Germany\\
{\tt\small Corresponding Author: hannan.khan@dfki.de}
}
\begin{document}
%
\maketitle
\begin{abstract}
With recent advances in computer vision, it appears that autonomous driving will be part of modern society sooner rather than later. However, there are still a significant number of concerns to address. Although modern computer vision techniques demonstrate superior performance, they tend to prioritize accuracy over efficiency, which is a crucial aspect of real-time applications. Large object detection models typically require higher computational power, which is achieved by using more sophisticated onboard hardware. For autonomous driving, these requirements translate to increased fuel costs and, ultimately, a reduction in mileage. Further, despite their computational demands, the existing object detectors are far from being real-time. In this research, we assess the robustness of our previously proposed, highly efficient pedestrian detector LSFM on well-established autonomous driving benchmarks, including diverse weather conditions and nighttime scenes. Moreover, we extend our LSFM model for general object detection to achieve real-time object detection in traffic scenes. We evaluate its performance, low latency, and generalizability on traffic object detection datasets. Furthermore, we discuss the inadequacy of the current key performance indicator employed by object detection systems in the context of autonomous driving and propose a more suitable alternative that incorporates real-time requirements.
\end{abstract}
\section{Introduction}
\label{sec:intro}

Autonomous driving aims to improve road safety, comfort, traffic congestion, and fuel consumption by replacing human drivers. The promise of autonomous driving is revolutionary, but it comes with many challenges. The pipeline of autonomous driving systems comprises numerous modules, with perception being the first. The primary function of the perception system is to obtain vital information from the surrounding environment of the ego vehicle and transmit it to the autonomous system in a readily consumable format. It is one of the most computationally demanding modules, as it works with raw data. The computational cost directly affects the mileage of the autonomous vehicle, as it directly translates to fuel costs and increases hardware requirements. A reasonable setup with a powerful GPU can alone cost significant mileage, while existing object detection approaches are far from real-time ($30\; FPS$). In addition to object detection, the perception module has multiple perception subroutines, which further tighten the constraints. Therefore, a lightweight object detector with superior accuracy, a minimal hardware footprint, and computational efficiency is desired.

\begin{figure}[!t]
\centering
\includegraphics[width=0.47\textwidth]{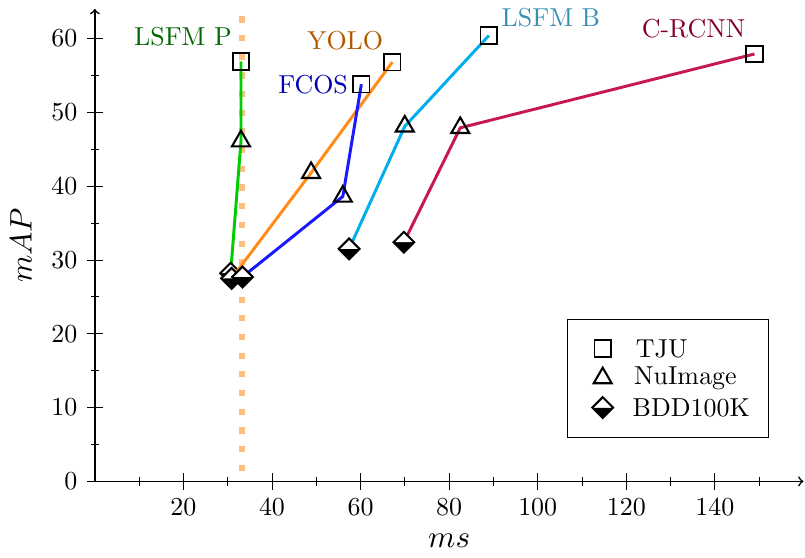}
\caption{
  Comparison of LSFM models with different traffic object detection models on real-world autonomous driving datasets. The dotted yellow line indicates the real-time threshold. LSFM P is the only model to achieve $30 FPS$ on all datasets with reasonable $mAP$.
  }
  \label{fig:vs}
\end{figure}

Object detection is one of the most crucial components of autonomous driving perception systems. The R-CNN \cite{girshick2014rich} is one of the first architectures of object detection with a reasonable level of accuracy, and it has proven effective in most applications. Nonetheless, its architecture indicates that it is a make-around solution for object detection, as the primary objective of R-CNNs is to extract regions of interest and pass them to an image classification network \cite{deng2009imagenet}. Cascade R-CNN \cite{cai2018cascade} is an R-CNN based architecture that improves performance by employing more sophisticated detection heads. However, they still suffer from the same inefficiency. Single-stage architectures \cite{yolo, redmon2018yolov3}, such as YOLO \cite{yolo}, try to solve the inefficiency of R-CNNs \cite{girshick2014rich} by replacing region proposal networks with predefined anchors. The approach is faster than two-stage approaches, but still searches the entire image for objects with predefined anchors. Furthermore, the performance of single-stage architecture is inferior compared to two-stage architecture. Recently, \textit{Vision Transformers}(ViT) \cite{dosovitskiy2020image} based solutions for object detection have demonstrated superior performance \cite{dyhead, detr, liu2021swin, gao2021fast}. However, these architectures come with inefficient and computationally costly components, specifically self-attention. Recent advances in anchor-free object detectors \cite{law2018cornernet, duan2019centernet, fcos, khan2022f2dnet} tend to bridge the gap between performance and efficiency and offer better trade-offs than anchor-based architectures. Anchor-free architectures detect objects in an end-to-end, per-pixel manner by formulating objects as pairs \cite{law2018cornernet} or triplets \cite{duan2019centernet} of keypoints. This formulation eliminates the need for anchor-based training and trains in an end-to-end fashion instead. Although anchor-free architectures are more performant than single-stage architectures, they still lag compared to two-stage and ViT-based architectures.

Furthermore, key performance indicators, or KPIs, provide a quantitative measure for assessing different approaches to a problem. The mean average precision, commonly known as $mAP$, is a well-recognized KPI for object detection. It involves the summation of precision-recall curves per class into average precision per class, and the mean of these values across all classes yields a singular value, i.e., $mAP$. The $mAP$ is a good KPI for object detection due to its ability to accommodate false alarms and missed objects, making it suitable for applications with different sensitivities. However, it lacks specificity for autonomous driving, as it does not incorporate the real-time critical requirements of autonomous driving. This raises questions regarding the suitability of object detectors with higher $mAP$ for real-time applications, and also reorients the research community in a manner that is not in line with the advancements in autonomous driving.

Pedestrians are crucial traffic objects from the perspective of autonomous driving, as a collision between a vehicle and a pedestrian can be deadly. Also, detecting pedestrians is harder due to their diverse clothing and apparent sizes. It is a prevalent practice within the research community to employ sophisticated object detection architectures for pedestrian detection. However, if an architecture performs well for pedestrian detection with additional constraints, it should perform well when extended to other traffic objects. Our recently proposed, LSFM \cite{khan2023localized} achieved the state-of-the-art performance in pedestrian detection. It is robust against motion blur, has a shorter inference time, and works well, especially in small and heavily occluded cases. With the goal of achieving real-time object detection, in this work, we extend LSFM to multiple classes and determine its generalizability to traffic object detection. We also evaluate its generalizability on synthetic datasets, and under severe weather and lighting conditions, including nighttime. Furthermore, we propose a precise key performance indicator for real-time object detection. Finally, we benchmark LSFM models across a diverse range of traffic object detection datasets, utilizing conventional and real-time evaluation metrics for object detection.

The major contributions of this work are as follows;
\begin{itemize}
    \item We evaluate the generalizability of LSFM \cite{khan2023localized} in night scenes and compare it on the KITTI \cite{Geiger2013IJRR} leaderboard.
    \item We extend the LSFM \cite{khan2023localized} by incorporating multi-class object detection to facilitate traffic object detection.
    \item We propose a novel key performance indicator for real-time object detection.
    \item We evaluate LSFM \cite{khan2023localized} for traffic object detection on well-established autonomous driving benchmarks, using conventional and real-time evaluation metrics.
\end{itemize}

\section{Related Work}
\label{sec:related_work}

Object detection aims to detect objects of interest in a given image. R-CNN \cite{girshick2014rich} is an early, deep learning based, two-staged object detection architecture. The idea of R-CNN is simple: use classification networks \cite{deng2009imagenet} to classify different parts of images or regions. Faster R-CNN \cite{fasterrcnn} proposed reusing convolutional features between regions. Cascade R-CNN \cite{cai2018cascade} proposed multiple detection heads to improve detection in a cascading manner. However, all R-CNN-based techniques are inherently inefficient with two-stage design, complex, and hence computationally expensive.


YOLO \cite{yolo} is a single-stage object detector which takes a simplified approach by dividing the image into a grid and predicting a fixed number of bounding boxes, confidence score, and classes per cell. Although fast, it has lower localization accuracy and performs poorly in small and crowded scenarios. The successor of YOLO, YOLOv3\cite{redmon2018yolov3} tries to improve performance while decreasing the inference time. SSD \cite{liu2016ssd} uses predefined bounding boxes of different scales and aspect ratios. It predicts confidence scores, bounding boxes deltas, and classes for each bounding box. SSD \cite{liu2016ssd} has lesser inference time than R-CNNs; however, the performance is worse. To bridge this performance gap, Retina Net\cite{focal_loss} introduces focal loss and argues that the gap is due to a foreground and background class imbalance.




Vision-Transformers or ViT \cite{dosovitskiy2020image} adapt transformer architecture from NLP for vision tasks. ViT-based networks are state-of-the-art in numerous vision tasks, including object detection \cite{liu2021swin, dyhead, detr}. ViT \cite{dosovitskiy2020image} splits images into $16\times16$ patches and treats them as tokens to feed into a transformers-based architecture. Swin transformers \cite{liu2021swin} propose sliding window-based tokenization to improve information flow between patches. Although ViT performs well on various tasks, they require enormous amounts of data and computational power to train and usually have longer inference times.


Anchor-free object detection approaches take the fixed grid idea of YOLO \cite{yolo} to another level by applying it on a per-pixel level, i.e., object probabilities are predicted per pixel, reducing the localization error, which YOLO \cite{yolo} like architecture are prone to. CornerNet \cite{law2018cornernet} presents the idea of detecting objects as paired keypoints. CenterNet \cite{duan2019centernet} models objects as keypoint triplets, introducing the center point to further refine detections, as the center point contains greater information about the object. FCOS \cite{fcos} takes a rather direct approach by detecting object centers and predicting bounding box dimensions as attributes of the center. Anchor-free approaches strike a good balance between efficiency and performance. However, they can be improved further as they use a basic CNN-based architecture.

\section{Efficient Traffic Object Detection}

LSFM \cite{khan2023localized} is an efficient object detector for pedestrian detection. Since pedestrians are the most challenging traffic objects, an efficient and highly performant pedestrian detection architecture should generalize well to other traffic objects. In this section, we first briefly explain the working of LSFM \cite{khan2023localized}, followed by its extension for traffic object detection, and finally propose a key performance indicator for object detection tailored for real-time scenarios like autonomous driving.

\subsection{Localized Semantic Feature Mixers}

LSFM \cite{khan2023localized} takes raw images as input and uses the ConvMLP-Pin backbone to extract high-level semantic features. These features are passed on to SP3, which splits them into patches of different sizes so that featuremaps from each stage produce an equal number of patches. Moreover, the patches corresponding to similar spatial locations are aligned, flattened, and concatenated to form a single $1D$ vector. They are passed through a single, fully connected layer to filter and enrich in a localized manner. Further, DFDN mixes these localized semantic features via MLPMixer blocks to detect objects; hence the name \textit{Localized Semantic Feature Mixers} \cite{khan2023localized}.

\subsection{Extension for Traffic Object Detection}
\label{sec:dfdn}
LSFM \cite{khan2023localized} uses high-level semantic feature representation of pedestrians, i.e., center, scale, and offset representation. $3$ objectives are formulated in the detection head, and each is optimized with a dedicated subnetwork. Binary cross entropy loss is used for center prediction with focal loss \cite{focal_loss} to make training robust to heavy background-foreground imbalance. Specifically, $\alpha$ variant of focal loss \cite{focal_loss} is used with $\alpha$ being a Gaussian base penalty reduction term to ease center learning. 






To extend the pedestrian detection model and enable multi-class object detection, the detection head needs to be changed to perform multi-class classification. Further, the scale and offset prediction branch can be left untouched as these attributes can be learned in a class-agnostic manner \cite{fcos}. For pedestrian detection, the loss is normalized by the number of object instances, this allows uniform focus on crowded as well as simpler scenarios during training. However, if the loss from all classes is simply accumulated and normalized with the total number of instances, the optimization will favor classes with higher density, i.e., cars in most cases. To solve this, we normalize center loss from each class separately by the number of occurrences in the batch. The final center loss equation for multiple objects becomes,

\begin{equation}
    \label{eq:focal_loss_c}
    L_{center} = \frac{1}{C} \sum_{c} \frac{1}{K_c} \sum_{t} \alpha_{c}(t) FL_{c}(p_{t}, y_{t}),
\end{equation}

where, $C$ and $K_c$ represent the number of classes and number of object instances in a class, while $\alpha_{c}$ and $FL_{c}$ are penalty reduction factor and focal loss similar to \cite{khan2023localized}, but for a particular class. We use similar loss weights as \cite{khan2023localized}.

\begin{figure}[!t]
\centering
\includegraphics[width=0.47\textwidth]{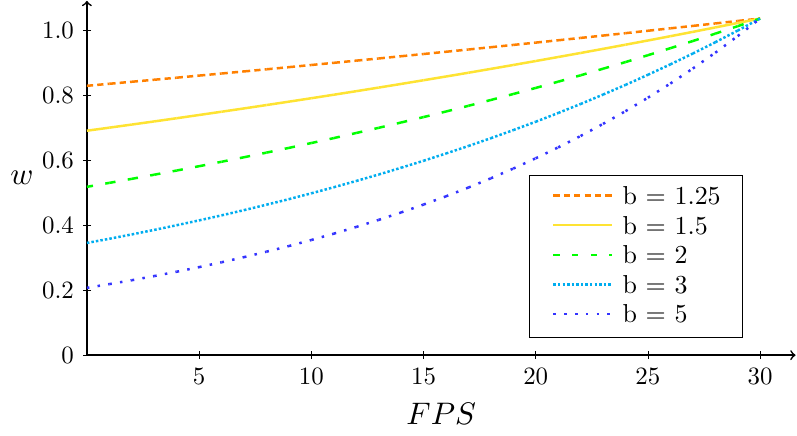}
\caption{The weighting factor $w$ of RTOP plotted against $FPS$ for different base $b$ values. Lower $b$ values increase the contribution of $p$ in RTOP, while higher values favor throughput.}
\label{fig:rtop}
\end{figure}

\subsection{Real-Time Objective Performance}
As autonomous driving requires time-critical perception, the perception tasks like object detection need to work in real-time. While the definition of real-time varies from domain to domain, $30 FPS$ is an acceptable threshold for autonomous driving case.

Mean average precision or $mAP$ is a well-known key performance indicator for object detection; however, it is independent of inference time and hence not suitable for real-time systems like autonomous driving. To this extent, we propose, Real-Time Objective Performance or $RTOP(mAP)$, which is a key performance indicator derived from $mAP$ for real-time systems. The following equation shows the relation for RTOP with performance $p$ and $FPS$.

\begin{equation}
\begin{aligned}
    \label{eq:rtop}
    RTOP_T(p, FPS) = \; &
        p \times w \quad \quad \quad \quad&,\\
        w = \; & b^{\phi - 1} &,\\
        \phi = \; & \min(\frac{FPS}{T}, 1) &,\\
\end{aligned}
\end{equation}

where, $p$ is performance measure, $mAP$ in our case, $T$ is real-time frame-rate, $b$ is the weight base which adjusts the scaling, and $\phi$ is frame-rate ratio. Fig. \ref{fig:rtop} shows the values of $w$ when using different $b$. We use $T = 30$ and $b = 2$ as these settings consider the performance and the real-time constraint equally.

\section{Results}

Before we begin the performance evaluation of the extended LSFM on traffic object detection, we first evaluate the impact of variable lighting conditions on the performance of LSFM. As no well-known, separate benchmark for object detection in night scenes exists, we evaluate LSFM on an existing pedestrian detection benchmark encompassing night scenes.


\subsection{Evaluation on KITTI Pedestrian Benchmark}

To ensure fair comparison, the test set of KITTI dataset \cite{Geiger2013IJRR} is withheld at the official server and evaluation of these sets is only possible by request at official server{\color{magenta}\footnote{\;\color{magenta}https://www.cvlibs.net/datasets/kitti/}}. Tab. \ref{tab:kitti_ped} show comparison of LSFM \cite{khan2023localized} on the leaderboard of KITTI \cite{Geiger2013IJRR}. LSFM \cite{khan2023localized} outperforms existing camera based published approaches with a significant margin, showing robustness to heavy occlusion. Inference time comparison is skipped, as other methods on the leaderboard do not provide detailed information about inference time and hardware used for testing. 

\begin{figure*}[!t]
    \centering
    \subfloat{\raisebox{2.6em}{\rotatebox[origin=t]{90}{{\small \textit{Cascade RCNN}}}}\;\:}
    \subfloat{\includegraphics[width=0.19\textwidth]{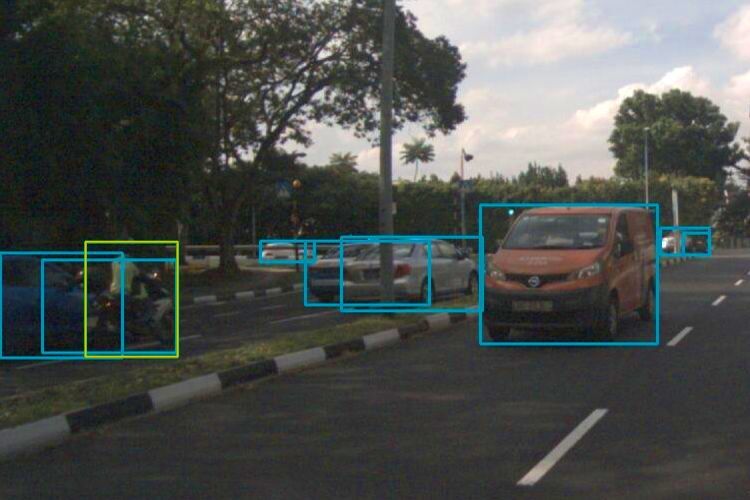}\hspace{0.004\textwidth}}
    \subfloat{\includegraphics[width=0.19\textwidth]{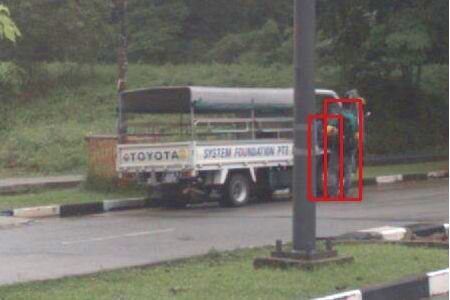}\hspace{0.004\textwidth}}
    \subfloat{\includegraphics[width=0.19\textwidth]{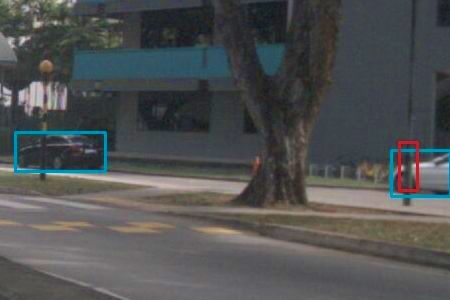}\hspace{0.004\textwidth}}
    \subfloat{\includegraphics[width=0.19\textwidth]{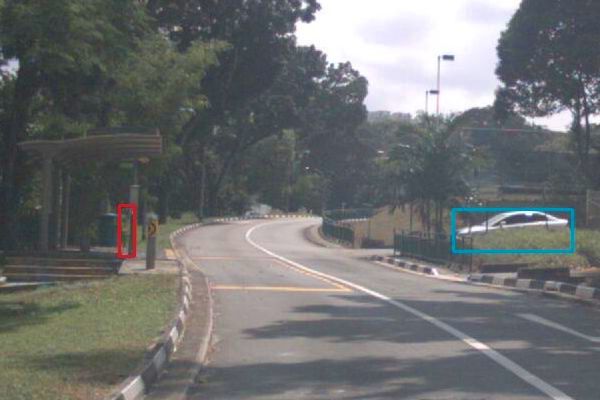}\hspace{0.004\textwidth}}
    \subfloat{\includegraphics[width=0.19\textwidth]{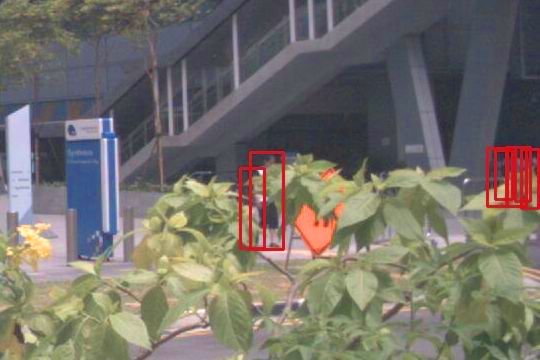}\hspace{0.004\textwidth}}
    \vspace{-0.9em}
    \subfloat{\raisebox{2.6em}{\rotatebox[origin=t]{90}{{\small \textit{Ground Truth}}}}\;\:}
    \subfloat{\includegraphics[width=0.19\textwidth]{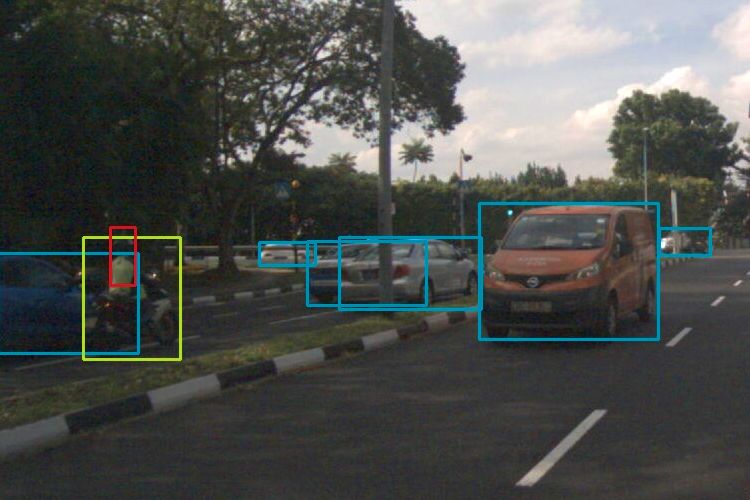}\hspace{0.004\textwidth}}
    \subfloat{\includegraphics[width=0.19\textwidth]{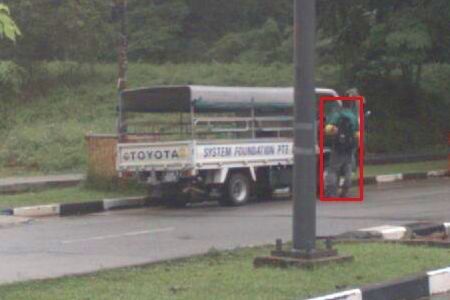}\hspace{0.004\textwidth}}
    \subfloat{\includegraphics[width=0.19\textwidth]{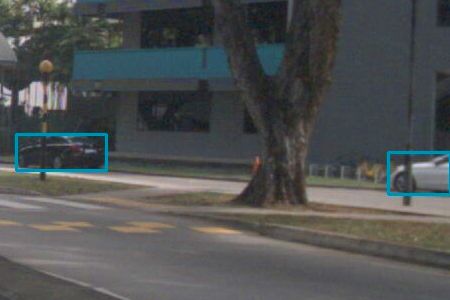}\hspace{0.004\textwidth}}
    \subfloat{\includegraphics[width=0.19\textwidth]{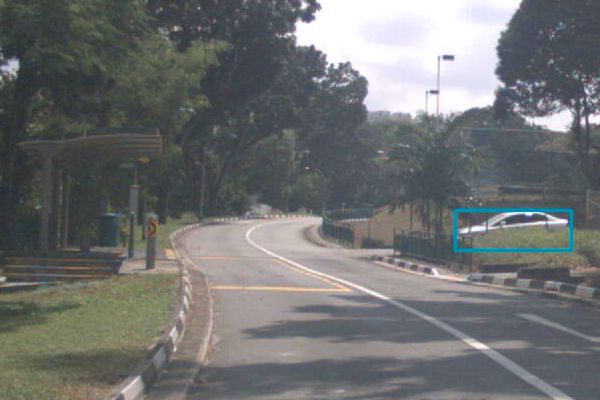}\hspace{0.004\textwidth}}
    \subfloat{\includegraphics[width=0.19\textwidth]{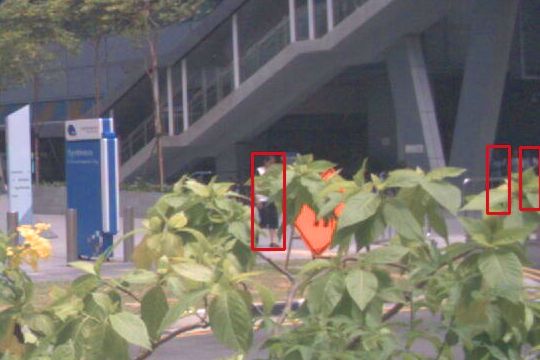}\hspace{0.004\textwidth}}
    \vspace{-0.9em}
    \subfloat{\raisebox{2.6em}{\rotatebox[origin=t]{90}{{\small \textit{LSFM}}}}\;\:}
    \subfloat{\includegraphics[width=0.19\textwidth]{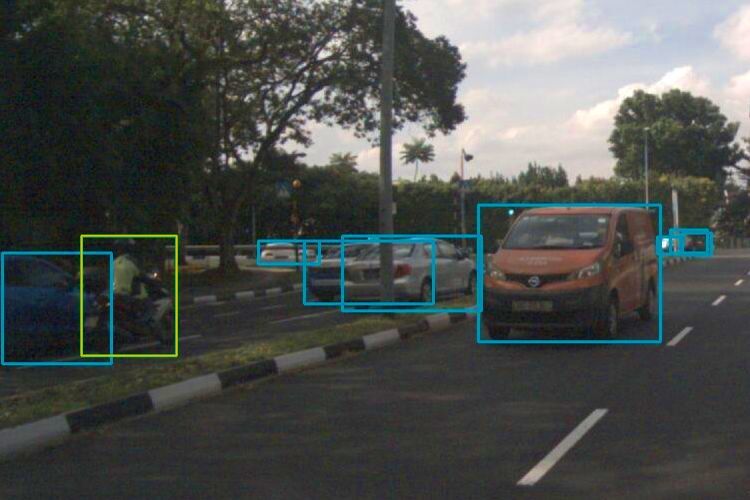}\hspace{0.004\textwidth}}
    \subfloat{\includegraphics[width=0.19\textwidth]{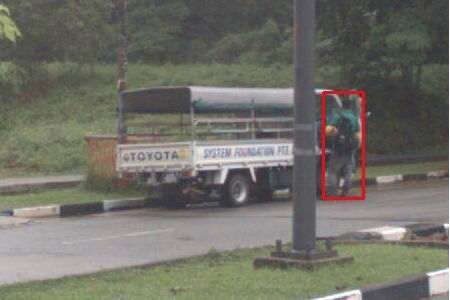}\hspace{0.004\textwidth}}
    \subfloat{\includegraphics[width=0.19\textwidth]{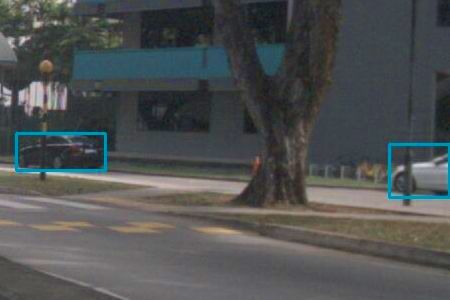}\hspace{0.004\textwidth}}
    \subfloat{\includegraphics[width=0.19\textwidth]{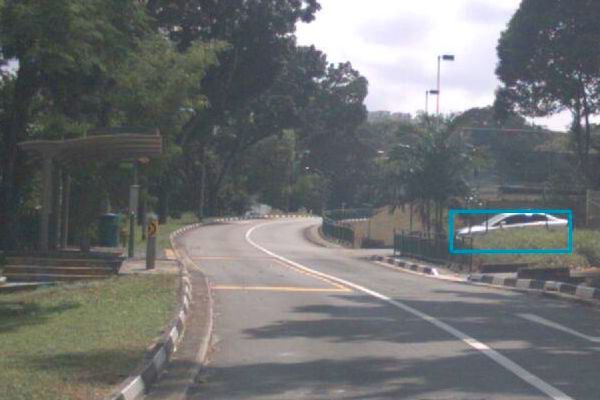}\hspace{0.004\textwidth}}
    \subfloat{\includegraphics[width=0.19\textwidth]{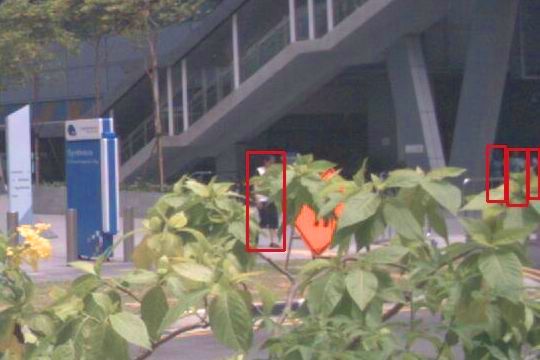}\hspace{0.004\textwidth}}
    \caption{Qualitative comparison of LSFM B and Cascade R-CNN \cite{cai2018cascade}. Car detections are indicated by the color cyan, pedestrian detections are indicated by the color red, and motorcycle detections are indicated by the color green. Other classes are ignored for simplicity in comparison. The contrast of the output images is enhanced for better visibility.}
    \label{fig:qual_res}
\end{figure*}

\begin{table}[!t]
    \renewcommand{\arraystretch}{1.3}
    \caption{LSFM \cite{khan2023localized} establishes a new state-of-the-art on the KITTI pedestrian benchmark \cite{Geiger2013IJRR}. For a fair comparison, only published and camera-based methods are listed.}
    \label{tab:kitti_ped}
    \centering
    \resizebox{0.9\linewidth}{!}{
        \begin{tabular}{l|ccc|c}
             \hline
             Method & $\uparrow$ Moderate & Easy & Hard & Mean \\
             \hline
             FFNet \cite{zhao2020monocular} & 75.8 & 87.2 & 69.9 & 77.6 \\
             MHN \cite{cao2019high} & 76.0 & 87.2 & 69.5 & 77.6 \\
             Aston-EAS \cite{wei2019enhanced}  & 76.1 & 86.7 & 70.6 & 77.8 \\
             Faster RCNN (ECP) \cite{eurocitypersons} & 76.3 & 86.0 & 70.6 & 77.6 \\
             RRC \cite{ren2017accurate} & 76.6 & 86.0 & 71.5 & 78.0 \\
             TuSimple \cite{yang2016exploit} & 78.4 & \textbf{88.9} & 73.7 & 80.3 \\
            \rowcolor{lgray}
            LSFM & \textbf{86.8 }& 81.3 & \textbf{77.6} & \textbf{81.9}\\
             \hline
        \end{tabular}
        }
\end{table}

\begin{table}[!t]
    \renewcommand{\arraystretch}{1.3}
    \caption{Comparison on the test set of the Euro City Persons dataset. LSFM \cite{khan2023localized} stands as second with slightly inferior performance compared to SPNet \cite{spnet}. * marks inference times calculated on Nvidia V100 GPU.}
    \label{tab:test_ecp}
    \centering
    \resizebox{0.95\linewidth}{!}{
        \begin{tabular}{l|ccc|c|c}
             \hline
             Method &$\downarrow$ Reas. & Small & Heavy & $mMR$ & Infe. \\
             \hline
             Faster R-CNN \cite{fasterrcnn} & 20.1 & 35.9 & 70.1 & 42.0 & - \\
             Pedestron \cite{pedestron} & 9.6 & 15.8 & 27.5 & 17.6 & 0.44s \\

            SPNet w FPN \cite{spnet} & 9.0 & 17.2 & 29.2 & 18.5 & 0.27s* \\
             
             Pedestrian2 & 7.1 & 12.7 & 24.4 & 14.7 & - \\
             SPNet w cascade \cite{spnet} & 6.6 & 11.9 & \textbf{23.1} & \textbf{13.9} & 0.27s* \\

             \rowcolor{lgray}
             LSFM & \textbf{6.3} & \textbf{11.1} & 25.3 & 14.2 & \textbf{0.17s} \\
             \hline
        \end{tabular}
        }
\end{table}

\begin{table}[!t]
    \renewcommand{\arraystretch}{1.3}
    \caption{Summary of the traffic object detection datasets. (*) the values are based on train and validation sets.}
    \label{tab:datasets_tod}
    \centering
    \resizebox{0.95\linewidth}{!}{
        \begin{tabular}{l|cccc}
             \hline
             Dataset & Images & Resolution & Objects & D/N \\
             \hline
             NuImages \cite{nuscenes} & 93K & 1600 $\times$ 900 & 800K & D/N 
              \\
            
             BDD100K \cite{yu2020bdd100k} & 100K & 1280 $\times$ 720 & 1800K & D/N \\
             
             Shift \cite{sun2022shift} & 2500K & 1280 $\times$ 800 & *1525K & D/N \\
            
             TJU-DHD-Traffic \cite{tju} & 60K & 1624 $\times$ 1200 & 332K & D/N \\
             \hline
        \end{tabular}
    }
\end{table}

\subsection{Performance at the Night Time}
Motion blur is one of the major factors causing localization inaccuracies for object detectors. As motion blur is caused due to changes in the scene while the camera shutter is open, it intensifies at the nighttime because of the increased open shutter duration. To evaluate the performance of LSFM \cite{khan2023localized} in extreme low lighting conditions (night) and how robust it is to intensified motion blur, we benchmark it on the Euro City Persons \cite{eurocitypersons} night dataset. Tab. \ref{tab:test_ecp} shows the performance of LSFM \cite{khan2023localized} on test set of Euro City Persons \cite{eurocitypersons}. LSFM \cite{khan2023localized} performs better than SPNet \cite{spnet} in reasonable and small cases at nighttime, but, the overall performance is slightly worse compared to SPNet \cite{spnet} with difference to $0.3\% mMR$. However, that performance gap between LSFM \cite{khan2023localized} and SPNet \cite{spnet} at nighttime is lesser compared to the gap at daytime, $0.8\% mMR$, which proves that LSFM \cite{khan2023localized} is robust to intense motion blur.

\subsection{Traffic Object Detection with LSFM}
Even though pedestrians pose a higher risk to autonomous driving, other road objects, such as cars, buses, barriers, traffic cones, and motorcycles, also require detection to avoid collision and drive safely. We take LSFM \cite{khan2023localized} and extend it for multi-class object detection to determine its scalability and generalizability. In this section, we first go through the traffic object detection datasets, followed by the comparison of LSFM models with the current state-of-the-art on them.

Over the past decade, a significant amount of research has been directed towards autonomous driving. One of the major achievements in this regard is the development of large-scale autonomous driving datasets \cite{caltech, Geiger2013IJRR}. The Caltech \cite{caltech} and KITTI \cite{Geiger2013IJRR} are early autonomous driving datasets; despite the lower number of samples and low resolution, these datasets contributed a lot to the development of autonomous driving. The NuImages dataset \cite{nuscenes}, released after the success of the NuScenes dataset \cite{nuscenes}, contains 2D object detection annotations belonging to $10$ different classes. The image resolution of NuImages \cite{nuscenes} is significantly higher than that of the KITTI dataset \cite{Geiger2013IJRR}, and it exhibits a greater diversity of environmental conditions. Moreover, it is richer in terms of object density, and the amount of data is good, containing $93K$ image samples. The recently released TJU-DHD dataset \cite{tju} has an even higher image resolution and also contains scenes from nighttime; however, it only contains $60K$ samples. The more recent BDD100K dataset \cite{yu2020bdd100k} has close to HD resolution, with $100K$ samples containing both day and night scenes in diverse weather conditions. Although it also contains objects of $10$ different classes, the labels are different from NuImages \cite{nuscenes}. Finally, Shift \cite{sun2022shift} is a synthetic autonomous driving dataset that was created to capture continuous domain shifts. The image resolution of the Shift dataset is similar to that of BDD100K \cite{yu2020bdd100k}; however, it comprises $2.5$ million images that capture diverse weather, lightning, and road conditions. Tab. \ref{tab:datasets_tod} contains a summary of these datasets.

\begin{table}[!t]
    \renewcommand{\arraystretch}{1.3}
    \caption{Comparison of LSFM \cite{khan2023localized} on traffic object detection benchmarks. * indicates the results on official benchmarks.
    }
    \label{tab:rtop}
    \centering
    \resizebox{\linewidth}{!}{
        \begin{tabular}{l|ccc|cc}
            \hline
            Method & mAP & mAP50 & mAP75 & $FPS$ & $\uparrow$ $RTOP_{30}(mAP)$ \\
            \hline
            \multicolumn{6}{c}{\textbf{TJU-DHD-Traffic} \cite{tju}}\\
            \hline
            *Cascade RCNN & 57.9 & 82.7 & 66.6 & 6.7 & 33.8
            \\
            \rowcolor{lgray}
            LSFM B & \textbf{60.4} & 85.7 & 70.0 & 11.2 & 39.1
            \\
            *FCOS & 53.8 & 80.0 & 60.1 & 16.6 & 39.5
            \\
            YOLOv3 & 56.8 & 85.4 & 64.1 & 14.9 & 40.1
            \\
            \rowcolor{lgray}
            LSFM P & 56.9 & 83.7 & 64.4 & 30.0 & \textbf{56.9}
            \\
            \hline
            \multicolumn{6}{c}{\textbf{NuImage} \cite{nuscenes}}\\
            \hline
            FCOS & 38.6 & 65.0 & 39.1 & 17.9 & 29.2
            \\
            Cascade RCNN & 47.9 & & & 12.1 & 31.7
            \\
            \rowcolor{lgray}
            LSFM B & \textbf{48.1} & 76.2 & 51.9 & 14.3 & 33.5
            \\
            YOLOv3 & 41.8 & 71.1 & 43.0 & 20.5 & 33.6
            \\
            \rowcolor{lgray}
            LSFM P & 46.1 & 74.6 & 48.7 & 30.3 & \textbf{46.1}
            \\
            \hline
            \multicolumn{6}{c}{\textbf{Shift} \cite{sun2022shift}}\\
            \hline
            Cascade RCNN & 48.6 & 64.1 & 52.8 & 13.9 & 33.5
            \\
            YOLOv3 & 45.9 & 69.1 & 48.6 & 23.4 & 39.4
            \\
            \rowcolor{lgray}
            LSFM B & \textbf{53.2} & 69.7 & 57.4 & 17.2 & 39.6
            \\
            FCOS & 46.2 & 63.9 & 48.9 & 27.0 & 43.1
            \\
            \rowcolor{lgray}
            LSFM P & 48.4 & 67.2 & 52.2 & 30.0 & \textbf{48.4}
            \\
            \hline
            \multicolumn{6}{c}{\textbf{BDD100K} \cite{yu2020bdd100k}}\\
            \hline
            *Cascade RCNN & \textbf{32.4} & & & 14.3 & 22.6
            \\
            \rowcolor{lgray}
            LSFM B & 31.5 & 59.1 & 29.0 & 17.4 & 23.6
            \\
            YOLOv3 & 27.5 & 54.5 & 23.8 & 32.4 & 27.5
            \\
            *FCOS & 27.7 & & & 30.0 & 27.7
            \\
            \rowcolor{lgray}
            LSFM P & 28.2 & 55.7 & 24.4 & 32.6 & \textbf{28.2}
            \\
            \hline
             
             \end{tabular}
    }
\end{table}

\subsubsection{Comparison with State-of-the-art}
To evaluate the performance of LSFM \cite{khan2023localized} for object detection, we compare it against existing architectures on well-known autonomous driving datasets. For an extensive comparison, we take multiple architectures of different kinds, i.e., we take anchor-based two-stage architecture (Cascade RCNN \cite{cai2018cascade}), anchor-based single-stage architecture (YOLOv3 \cite{redmon2018yolov3}), and anchor-free single-stage architecture (FCOS \cite{fcos}). We present results of two variants of LSFM, i.e., LSFM B and LSFM P, where LSFM B is the performant model with HRNet backbone while LSFM P is for real-time performance and features ConvMLP-Pin backbone \cite{khan2023localized}. To fairly compare the performance of LSFM \cite{khan2023localized} with other object detectors, we train all architectures without hard mixup augmentation.

Tab. \ref{tab:rtop} shows the comparison of LSFM models with the state-of-the-art object detectors. LSFM B outperforms the state-of-the-art on most datasets by a significant margin. On average, LSFM B performs $1.6\% mAP$ better than Cascade RCNN, $2.9\% mAP$ better than LSFM P, $5.3\% mAP$ better than YOLOv3 and $6.7\% mAP$ better than FCOS. Also, LSFM achieves $27\%$ lesser inference time compared to Cascade R-CNN. Although LSFM B has a higher inference time compared to FCOS and YOLOv3, there is a huge gap between their performance and the performance of LSFM B, with LSFM B leading the comparison. Further, LSFM P, which is an even more efficient model, achieves the least of all inference times, with on average $54\%$ lesser inference time compared to LSFM B. Also, with lesser inference time, LSFM P performs $1.9\% mAP$ and $3.3\% mAP$ better compared to YOLOv3 and FCOS respectively. However, LSFM P on average performs $1.3\% mAP$ worse compared to Cascade RCNN, but with only $\frac{1}{3}$ of its inference time.

\subsubsection{Real-Time Objective Performance} 

Given that certain models exhibit superior performance while others exhibit better inference time, it can be challenging to select the optimal model for real-time applications. Fig. \ref{fig:vs} shows the comparison of LSFM models with state-of-the-art based on performance and run-time. To ease the choice of the best model for real-time applications, we compare top-performing models in real-time settings using our proposed KPI, i.e., \textit{Real-Time Object Performance}. Tab. \ref{tab:rtop} shows the comparison of LSFM models on autonomous driving benchmarks in real-time settings. LSFM P outperforms existing methods by a significant margin, which implies that LSFM P is performant and well-suited for real-time systems. However, LSFM B performs better than Cascade RCNN but worse than the rest of the methods. This indicates that it is better suited for real-time applications than Cascade RCNN but worse than the rest. 

\subsubsection{Qualitative Comparison}
We qualitatively compare top-performing models, i.e., LSFM B and Cascade R-CNN, to analyze the visual difference between their detection. Fig. \ref{fig:qual_res} shows the qualitative comparison between LSFM B and Cascade R-CNN on the NuImages \cite{nuscenes} dataset. For this comparison, the confidence threshold is set to $0.3$, and only car, pedestrian, and motorcycle classes are selected to keep the comparison simple. The presented results only include the images where LSFM and Cascade R-CNN deviate, as most of the results from both models are similar. It is evident that Cascade R-CNN produces more false positives compared to LSFM B, especially in crowded scenes.

\section{Conclusion}

This paper adopts an unconventional approach by extending a well-established pedestrian detection architecture to detect multi-class objects. It asserts that detection architectures capable of addressing problems with more constraints, such as pedestrian detection, can handle multi-class object detection. To this extent, the paper evaluates LSFM in low lighting conditions and against a popular pedestrian detection leaderboard to establish its robustness and extend it for multi-class object detection. Further, it compares LSFM models with modern object detection architectures on well-established autonomous driving benchmarks. In most cases, LSFM B beats conventional object detection models significantly. The paper further argues that $mAP$ is insufficient for real-time object detection and proposes a novel KPI, $RTOP$, which fulfills this requirement. In comparison with modern object detectors in real-time settings, using $RTOP$ as an evaluation metric, LSFM P, a lighter and more efficient version of LSFM, beats the rest of the models by a significant margin, demonstrating its suitability for real-time applications such as autonomous driving.


{
\small
\bibliographystyle{IEEEbib}
\bibliography{root}
}

\end{document}